\documentclass[11pt]{article}
\usepackage{acl2000}
\usepackage[T1]{fontenc} %To get 'i' double dot to come out correctly
\title{Using existing systems to supplement small amounts of annotated
grammatical relations training data \thanks{\hspace{1em}This paper
reports on work performed at the MITRE Corporation under the support
of the MITRE Sponsored Research Program. Marc Vilain provided the
motivation to find GRs. Warren Greiff suggested using
randomization-type techniques to determine statistical
significance. Sabine Buchholz and John Carroll ran their GR finding
systems over our data for the experiments. Jun Wu provided some
helpful explanations. Christine Doran and John Henderson provided
helpful editing. Three anonymous reviewers provided helpful
suggestions.}  }

\author{Alexander Yeh\\Mitre Corp.\\202 Burlington Rd.\\Bedford, MA 01730\\USA\\asy@mitre.org}

\begin{document}

\maketitle

%\typeout{REMOVE PAGE NUMBERS IN FINAL PAPER!!!}

\begin{abstract}
{\begin{picture}(0,0)
\put(-20,215){Appears in the 38th Annual Meeting of the Association for Computational Linguistics (ACL-2000).}
\put(-10,200){pages 126-132, Hong Kong, October, 2000. Copyright \copyright 2000
Association for Computational Linguistics.}
\put(-20,-515){cs.CL/0010020}
\end{picture}}
Grammatical relationships (GRs) form an important level of natural
language processing, but different sets of GRs are useful for
different purposes. Therefore, one may often only have time to obtain
a small training corpus with the desired GR annotations. To boost the
performance from using such a small training corpus on a
transformation rule learner, we use existing systems that find related
types of annotations.
%We find that the improvement tends to concentrate on a few types of relations, with different existing systems concentrating on different types.  We use this tendency to further boost performance by merging the output of these existing systems.
\end{abstract}

\section{Introduction}

Grammatical relationships (GRs), which include arguments (e.g.,
subject and object) and modifiers, form an important level of natural
language processing. Examples of GRs in the sentence 
\begin{center}
{\em Today, my dog pushed the ball on the floor.} 
\end{center}
are {\em pushed} having the subject \mbox{\em my dog}, the object
\mbox{\em the ball} and the time modifier {\em Today}, and
\mbox{\em the ball} having the location modifier \mbox{\em on (the
floor)}. The resulting annotation is
\begin{center}
\mbox{\em my dog} $-${\tt subj}$\rightarrow$ {\em pushed}\\
{\em on} $-${\tt mod-loc}$\rightarrow$ \mbox{\em the ball}
\end{center}
etc. GRs are the objects of study in relational grammar
\cite{Perlmutter83}. In the SPARKLE project \cite{Sparkle1}, GRs form
the top layer of a three layer syntax scheme. Many systems (e.g., the
KERNEL system \cite{PPWF93}) use GRs as an intermediate form when
determining the semantics of syntactically parsed text.  GRs are often
stored in structures similar to the F-structures of lexical-functional
grammar \cite{Kaplan94}.

A complication is that different sets of GRs are useful for different
purposes. For example, \newcite{FVY99} is interested in semantic
interpretation, and needs to differentiate between time, location and
other modifiers. The SPARKLE project \cite{Sparkle1}, on the other
hand, does not differentiate between these types of modifiers. As has
been mentioned by John Carroll (personal communication), combining
modifier types together is fine for information retrieval. Also, having
less differentiation of the modifiers can make it easier to find them
\cite{FVY99}.

Furthermore, unless the desired set of GRs matches the set already
annotated in some large training corpus,\footnote{One example is a memory-based GR
finder \cite{BVD99} that uses the GRs annotated in the Penn Treebank
\cite{PennTreebank}.} one will have to either manually write rules to
find the GRs, as done in \newcite{AandC97}, or annotate a new
training corpus for the desired set. Manually writing rules is
expensive, as is annotating a large corpus.

Often, one may only have the resources to produce a small annotated
training set, and many of the less common features of the set's domain
may not appear at all in that set. In contrast are existing systems
that perform well (probably due to a large annotated training set or a
set of carefully hand-crafted rules) on related (but different)
annotation standards. Such systems will cover many more domain
features, but because the annotation standards are slightly different,
some of those features will be annotated in a different way than in
the small training and test set.

A way to try to combine the different advantages of these small
training data sets and existing systems which produce related
annotations is to use a sequence of two systems. We first use an
existing annotation system which can handle many of the less common
features, i.e., those which do not appear in the small training set. We
then train a second system with that same small training set to take
the output of the first system and correct for the differences in
annotations. This approach was used by \newcite{Palmer97} for word
segmentation. \newcite{Hwa99} describes a somewhat similar approach
for finding parse brackets which combines a fully annotated related
training data set and a large but incompletely annotated final
training data set. Both these works deal with just one (word boundary)
or two (start and end parse bracket) annotation label types and the
same label types are used in both the existing annotation
system/training set and the final (small) training set. In comparison,
our work handles many annotation label types, and the translation from
the types used in the existing annotation system to the types in the
small training set tends to be both more complicated and most easily
determined by empirical means. Also, the type of baseline score being
improved upon is different. Our work adds an existing system to
improve the rules learned, while \newcite{Palmer97} adds rules to
improve an existing system's performance.

We use this related system/small training set combination to improve
the performance of the transformation-based error-driven learner
described in \newcite{FVY99}.  So far, this learner has started with a
blank initial labeling of the GRs. This paper describes experiments
where we replace this blank initial labeling with the output from an
existing GR finder that is good at a somewhat different set of GR
annotations. With each of the two existing GR finders that we use, we
obtained improved results, with the improvement being more noticeable
when the training set is smaller.

We also find that the existing GR finders are quite uneven on how they
improve the results. They each tend to concentrate on improving the
recovery of a few kinds of relations, leaving most of the other kinds
alone. 

We use this tendency to further boost the learner's performance by
using a merger of these existing GR finders' output as the initial
labeling.

\section{The Experiment}

We now improve the performance of the \newcite{FVY99}
transformation rule learner on a small annotated training set by using
an existing system to provide initial GR annotations. This experiment
is repeated on two different existing systems, which are reported in
\newcite{BVD99} and \newcite{CMB99}, respectively.

Both of these systems find a somewhat different set of GR annotations
than the one learned by the \newcite{FVY99} system. For example, the
\newcite{BVD99} system ignores verb complements of verbs and is
designed to look for relationships to verbs and not GRs that exist
between nouns, etc. This system also handles relative clauses
differently. For example, in \mbox{\em ``Miller, who organized ...''},
this system is trained to indicate that {\em ``who''} is the subject
of {\em ``organized''}, while the \newcite{FVY99} system is trained to
indicate that {\em ``Miller''} is the subject of {\em
``organized''}. As for the \newcite{CMB99} system, among other things,
it does not distinguish between sub-types of modifiers such as time,
location and possessive. Also, both systems handle copulas (usually
using the verb ``to be'') differently than in \newcite{FVY99}.

\subsection{Experiment Set-Up}

\begin{figure*}
\setlength{\unitlength}{0.00083300in}%
\begingroup\makeatletter\ifx\SetFigFont\undefined
% extract first six characters in \fmtname
\def\x#1#2#3#4#5#6#7\relax{\def\x{#1#2#3#4#5#6}}%
\expandafter\x\fmtname xxxxxx\relax \def\y{splain}%
\ifx\x\y   % LaTeX or SliTeX?
\gdef\SetFigFont#1#2#3{%
  \ifnum #1<17\tiny\else \ifnum #1<20\small\else
  \ifnum #1<24\normalsize\else \ifnum #1<29\large\else
  \ifnum #1<34\Large\else \ifnum #1<41\LARGE\else
     \huge\fi\fi\fi\fi\fi\fi
  \csname #3\endcsname}%
\else
\gdef\SetFigFont#1#2#3{\begingroup
  \count@#1\relax \ifnum 25<\count@\count@25\fi
  \def\x{\endgroup\@setsize\SetFigFont{#2pt}}%
  \expandafter\x
    \csname \romannumeral\the\count@ pt\expandafter\endcsname
    \csname @\romannumeral\the\count@ pt\endcsname
  \csname #3\endcsname}%
\fi
\fi\endgroup
\begin{picture}(7524,2647)(140,-1871)
\thicklines
\put(2287,-511){\oval(1572,600)}
\put(2290,-589){\makebox(0,0)[b]{\smash{\SetFigFont{10}{12.0}{rm}existing system}}}
\put(1726,-436){\makebox(0,0)[b]{\smash{\SetFigFont{12}{14.4}{rm}+}}}
\put(2288,-1262){\oval(1572,600)}
\put(2291,-1340){\makebox(0,0)[b]{\smash{\SetFigFont{10}{12.0}{rm}existing system}}}
\put(1727,-1187){\makebox(0,0)[b]{\smash{\SetFigFont{12}{14.4}{rm}+}}}
\put(303,-1713){\framebox(750,300){}}
\put(678,-1638){\makebox(0,0)[b]{\smash{\SetFigFont{10}{12.0}{rm}test set}}}
\put(5851,164){\oval(1050,450)}
\put(5853,-663){\circle{474}}
\put(5778,-1638){\oval(1200,450)}
\put(5851,-61){\vector( 0,-1){375}}
\put(1576, 14){\vector( 1, 0){3750}}
\put(3376,539){\line( 1, 0){1500}}
\put(4876,539){\vector( 3,-2){450}}
\put(152,-212){\framebox(1425,375){}}
\put(901,-211){\vector( 2,-1){600}}
\put(1028,-1451){\vector( 2, 1){450}}
\put(4876,-1261){\vector( 4,-1){600}}
\put(226,314){\dashbox{4}(3150,450){}}
\put(4898,-514){\vector( 4, 3){600}}
\put(5852,-887){\vector( 0,-1){525}}
\put(3077,-1262){\vector( 1, 0){450}}
\put(3529,-1564){\dashbox{4}(1350,600){}}
\put(3077,-512){\vector( 1, 0){450}}
\put(3528,-813){\dashbox{4}(1350,675){}}
\put(6452,-1337){\dashbox{4}(1200,600){}}
\put(6413,-1647){\vector( 2, 1){600}}
\put(1053,-1711){\vector( 1, 0){4125}}
\put(827,-62){\makebox(0,0)[b]{\smash{\SetFigFont{10}{12.0}{rm}small training set}}}
\put(5852, 88){\makebox(0,0)[b]{\smash{\SetFigFont{10}{12.0}{rm}rule learner}}}
\put(1801,464){\makebox(0,0)[b]{\smash{\SetFigFont{10}{12.0}{rm}key GR annotations for small training set}}}
\put(5251,-436){\makebox(0,0)[b]{\smash{\SetFigFont{12}{14.4}{rm}*}}}
\put(5251,-1336){\makebox(0,0)[b]{\smash{\SetFigFont{12}{14.4}{rm}*}}}
\put(5853,-738){\makebox(0,0)[b]{\smash{\SetFigFont{10}{12.0}{rm}rules}}}
\put(3601,-361){\makebox(0,0)[b]{\smash{\SetFigFont{12}{14.4}{rm}+}}}
\put(4205,-1478){\makebox(0,0)[b]{\smash{\SetFigFont{10}{12.0}{rm}GR annotations}}}
\put(4206,-1191){\makebox(0,0)[b]{\smash{\SetFigFont{10}{12.0}{rm}initial test}}}
\put(3603,-1113){\makebox(0,0)[b]{\smash{\SetFigFont{12}{14.4}{rm}+}}}
\put(4205,-365){\makebox(0,0)[b]{\smash{\SetFigFont{10}{12.0}{rm}initial training}}}
\put(4205,-653){\makebox(0,0)[b]{\smash{\SetFigFont{10}{12.0}{rm}GR annotations}}}
\put(7053,-963){\makebox(0,0)[b]{\smash{\SetFigFont{10}{12.0}{rm}final test}}}
\put(7053,-1251){\makebox(0,0)[b]{\smash{\SetFigFont{10}{12.0}{rm}GR annotations}}}
\put(5779,-1714){\makebox(0,0)[b]{\smash{\SetFigFont{10}{12.0}{rm}rule interpreter}}}
\end{picture}
\caption{Set-up to use an existing system to improve performance}\label{f:after-help}
\end{figure*}
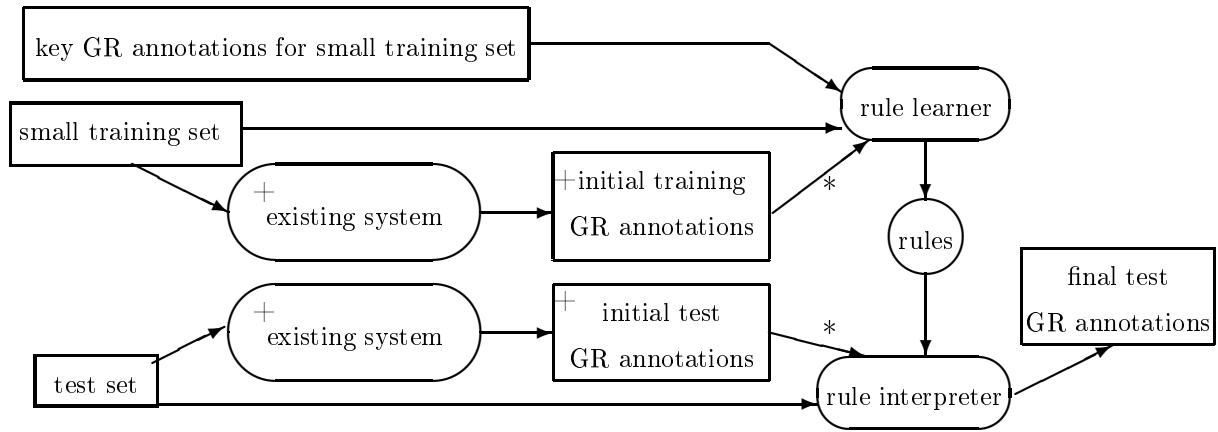

As described in \newcite{FVY99}, the transformation rule learner
starts with a p-o-s tagged corpus that has been ``chunked'' into noun
chunks, etc. The starting state also includes imperfect estimates of
pp-attachments and a blank set of initial GR annotations.  In these
experiments, this blank initial set is changed to be a translated
version of the annotations produced by an existing system. This is how
the existing system transmits what it found to the rule learner. The
set-up for this experiment is shown in figure~\ref{f:after-help}. The
four components with + signs are taken out when one wants the
transformation rule learner to start with a blank set of initial GR
annotations.

The two arcs in that figure with a * indicate where the translations
occur.  These translations of the annotations produced by the existing
system are basically just an attempt to map each type of annotation
that it produces to the most likely type of corresponding annotation
used in the \newcite{FVY99} system. For example, in our experiments,
the \newcite{BVD99} system uses the annotation {\tt np-sbj} to
indicate a subject, while the \newcite{FVY99} system uses the
annotation {\tt subj}. We create the mapping by examining the training
set to be given to the \newcite{FVY99} system. For each type of
relation $e_i$ output by the existing system when given the training
set text, we look at what relation types (which $t_k$'s) co-occur with
$e_i$ in the training set. We look at the $t_k$'s with the highest
number of co-occurrences with that $e_i$.  If that $t_k$ is unique (no
ties for the highest number of co-occurrences) and translating $e_i$
to that $t_k$ generates at least as many correct annotations in the
training set as false alarms, then make that translation.  Otherwise,
translate $e_i$ to no relation. This latter translation is not
uncommon. For example, in one run of our experiments, 9\% of the
relation instances in the training set were so translated, in another
run, 46\% of the instances were so translated.

Some relations in the \newcite{CMB99} system are between three or four
elements. These relations are each first translated into a set of two
element sub-relations before the examination process above is performed.

Even before applying the rules, the translations find many of the desired
annotations. However, the rules can considerably improve what is
found. For example, in two of our early experiments, the translations
by themselves produced F-scores (explained below) of about 40\% to
50\%. After the learned rules were applied, those F-scores increased
to about 70\%.

An alternative to performing translations is to use the {\em
un\/}translated initial annotations as an additional type of input to
the rule system. This alternative, which we have yet to try, has the
advantage of fitting into the transformation-based error-driven paradigm
\cite{BandR94} more cleanly than having a translation stage. However,
this additional type of input will also further slow-down an already
slow rule-learning module.

\subsection{Overall Results} \label{ss:overall-results}

For our experiment, we use the same 1151 word (748 GR) test set
used in \newcite{FVY99}, but for a training set, we use only a subset of
the 3299 word training set used in \newcite{FVY99}. This subset contains
1391 (71\%) of the 1963 GR instances in the original training set. The
overall results for the test set are
\begin{center}
\begin{tabular}{|l|r|r|r|r|}\hline
\multicolumn{5}{|c|}{Smaller Training Set, Overall Results}\\
 & R & P & F & ER\\ \hline
IaC & {\em 478 (63.9\%)} & 77.2\% & {\em 69.9\%} & 7.7\% \\ \hline
IaB & {\em 466 (62.3\%)} & 78.1\% & {\em 69.3\%} & 5.8\% \\ \hline
NI &  448 (59.9\%) & 77.1\% & 67.4\% & \\ \hline
\end{tabular}
\end{center}
where row~IaB is the result of using the rules learned when the
\newcite{BVD99} system's translated GR annotations are used as the
Initial Annotations, row~IaC is the similar result with the
\newcite{CMB99} system, and row~NI is the result of using the rules
learned when No Initial GR annotations are used (the rule learner as
run in \newcite{FVY99}).  R(ecall) is the number (and percentage) of
the keys that are recalled. P(recision) is the number of correctly
recalled keys divided by the number of GRs the system claims to
exist. \mbox{F(-score)} is the harmonic mean of recall ($r$) and precision
($p$) percentages. It equals $2pr/(p+r)$. ER stands for Error
Reduction. It indicates how much adding the initial annotations
reduced the missing \mbox{F-score}, where the missing \mbox{F-score}
is \mbox{100\%$-$F}. ER=
\mbox{100\%$\times$(F$_{IA}-$F$_{NI}$)/(100\%$-$F$_{NI}$)}, where
F$_{NI}$ is the F-score for the NI row, and F$_{IA}$ is the F-score
for using the Initial Annotations of interest. Here, the differences
in recall and F-score between NI and either IaB or IaC (but not
between IaB and IaC) are statistically significant. The differences in
precision is not.\footnote{When comparing differences in this paper,
the statistical significance of the higher score being better than the
lower score is tested with a one-sided test. Differences deemed
statistically significant are significant at the 5\% level.
Differences deemed non-statistically significant are not significant
at the 10\% level. For recall, we use a sign test for matched-pairs
\cite[Sec.~15.5]{Harnett82}. For precision and F-score, a
``matched-pairs'' randomization test \cite[Sec.~5.3]{Cohen95} is
used.} In these results, most of the modest F-score gain came from
increasing recall.

One may note that the error reductions here are smaller than
\newcite{Palmer97}'s error reductions. Besides being for different
tasks (word segmentation versus GRs), the reductions are also computed
using a different type of baseline. In \newcite{Palmer97}, the
baseline is how well an existing system performs before the rules are
run. In this paper, the baseline is the performance of the rules
learned without first using an existing system. If we were to use the
same baseline as \newcite{Palmer97}, our baseline would be an F of
37.5\% for IaB and 52.6\% for IaC. This would result in a much higher
ER of 51\% and 36\%, respectively.

We now repeat our experiment with the full 1963 GR instance training
set. These results indicate that as a small training set gets larger,
the overall results get better and the initial annotations help less
in improving the overall results. So the initial annotations are more
helpful with smaller training sets. The overall results on the test
set are
\begin{center}
\begin{tabular}{|l|r|r|r|r|}\hline
\multicolumn{5}{|c|}{Full Training Set, Overall Results}\\
 & R & P & F & ER \\ \hline
IaC & {\em 487 (65.1\%)} & {\em 79.7\%} & {\em 71.7\%} & 6.3\%\\ \hline
IaB & 486 (65.0\%) & 76.5\% & 70.3\% & 1.7\% \\ \hline
NI  &  476 (63.6\%) & 77.3\% & 69.8\% &\\ \hline
\end{tabular}
\end{center}
The differences in recall, etc. between IaB and NI are now small
enough to be not statistically significant. The differences between
IaC and NI are statistically significant,\footnote{The recall
difference is semi-significant, being significant at the 10\% level.}
but the difference in both the absolute F-score (1.9\% versus 2.5\%
with the smaller training set) and ER (6.3\% versus 7.7\%) has
decreased.

\subsection{Results by Relation}

The overall result of using an existing system is a modest increase
in F-score. However, this increase is quite unevenly distributed, with
a few relation(s) having a large increase, and most relations not
having much of a change. Different existing systems seem to have
different relations where most of the increase occurs.

As an example, take the results of using the \newcite{BVD99} system on
the 1391 GR instance training set.  Many GRs, like {\em possessive
modifier}, are not affected by the added initial annotations. Some
GRs, like {\em location modifier}, do slightly better (as measured by
the F-score) with the added initial annotations, but some, like {\em
subject}, do better without. With GRs like {\em subject}, some
differences between the initial and desired annotations may be too
subtle for the \newcite{FVY99} system to adjust for. Or those
differences may be just due to chance, as the result differences in
those GRs are not statistically significant. The GRs with
statistically significant result differences are the {\em time} and
``other''\footnote{Modifiers that do not fall into any of the subtypes
used, such as time, location, possessive, etc. Examples of {\em
unused} subtypes are purpose and modality.} {\em modifier}s, where
adding the initial annotations helps. The {\em time
modifier}\footnote{There are 45 instances in the test set key.}
results are quite different:
\begin{center}
\begin{tabular}{|l|r|r|r|r|}\hline
\multicolumn{5}{|c|}{Smaller Training Set, {\em Time Modifier}\/s}\\
& R & P & F & ER\\ \hline
IaB & {\em 29 (64.4\%)} & {\em 80.6\%} & {\em 71.6\%} & 53\% \\ \hline
NI &  14 (31.1\%) & 56.0\% & 40.0\% & \\ \hline
\end{tabular}
\end{center}
The difference in the number recalled (15) for this GR accounts for
nearly the entire difference in the overall recall results (18). The
recall, precision and F-score differences are all statistically
significant.

Similarly, when using the \newcite{CMB99} system on this training set,
most GRs are not affected, while others do slightly better. The only
GR with a statistically significant result difference is {\em object},
where again adding the initial annotations helps:
\begin{center}
\begin{tabular}{|l|r|r|r|r|}\hline
\multicolumn{5}{|c|}{Smaller Training Set, {\em Object} Relations}\\
& R & P & F & ER\\ \hline
IaC & {\em 198 (79.5\%)} & 79.5\% & {\em 79.5\%} & 17\% \\ \hline
NI &  179 (71.9\%) & 78.9\% & 75.2\% & \\ \hline
\end{tabular}
\end{center}
The difference in the number recalled (19) for this GR again accounts
for most of the difference in the overall recall results (30). The
recall and F-score differences are statistically significant. The
precision difference is not.

As one changes from the smaller 1391 GR instance training set to the
larger 1963 GR instance training set, these F-score improvements
become smaller. When using the \newcite{BVD99} system, the improvement
in the ``other'' {\em modifier} is now no longer statistically
significant. However, the {\em time modifier} F-score improvement
stays statistically significant:
\begin{center}
\begin{tabular}{|l|r|r|r|r|}\hline
\multicolumn{5}{|c|}{Full Training Set, {\em Time Modifier}\/s}\\
& R & P & F & ER\\ \hline
IaB & {\em 29 (64.4\%)} & {\em 74.4\%} & {\em 69.0\%} & 46\% \\ \hline
NI &  15 (33.3\%) & 57.7\% & 42.3\% & \\ \hline
\end{tabular}
\end{center}
When using the \newcite{CMB99} system, the {\em object} F-score
improvement stays statistically significant:
\begin{center}
\begin{tabular}{|l|r|r|r|r|}\hline
\multicolumn{5}{|c|}{Full Training Set, {\em Object} Relations}\\
& R & P & F & ER\\ \hline
IaC & 194 (77.9\%) & {\em 85.1\%} & {\em 81.3\%} & 16\% \\ \hline
NI &  188 (75.5\%) & 80.3\% & 77.8\% & \\ \hline
\end{tabular}
\end{center}

\subsection{Combining Sets of Initial Annotations}

So the initial annotations from different existing systems tend to
each concentrate on improving the performance of different GR
types. From this observation, one may wonder about combining the
annotations from these different systems in order to increase the
performance on all the GR types affected by those different existing
systems.

Various works \cite{HZD98,HandB99,WandS98} on combining different
systems exist. These works use one or both of two types of
schemes. One is to have the different systems simply
vote. However, this does not really make use of the fact that
different systems are better at handling different GR types.  The
other approach uses a combiner that takes the systems' output as input and
may perform such actions as determining which system to use under
which circumstance. Unfortunately, this approach needs extra training data
to train such a combiner. Such data may be more useful when used
instead as additional training data for the individual methods that
one is considering to combine, especially when the systems being
combined were originally given a small amount of training data.

To avoid the disadvantages of these existing schemes, we came up with
a third method.  We combine the existing related systems by
taking a union of their translated annotations as the new initial GR
annotation for our system. We rerun rule learning on the smaller (1391
GR instance) training set with a Union of the \newcite{BVD99} and
\newcite{CMB99} systems' translated GR annotations. The overall
results for the test set are (shown in row IaU)
\begin{center}
\begin{tabular}{|l|r|r|r|r|}\hline
\multicolumn{5}{|c|}{Smaller Training Set, Overall Results}\\
 & R & P & F & ER\\ \hline
IaU & {\em 496 (66.3\%)} & 76.4\% & {\em 71.0\%} & 11\%  \\ \hline
IaC & {\em 478 (63.9\%)} & 77.2\% & {\em 69.9\%} & 7.7\% \\ \hline
IaB & {\em 466 (62.3\%)} & 78.1\% & {\em 69.3\%} & 5.8\% \\ \hline
NI &  448 (59.9\%) & 77.1\% & 67.4\% & \\ \hline
\end{tabular}
\end{center}
where the other rows are as shown in
Section~\ref{ss:overall-results}. Compared to the F-score with using
\newcite{CMB99} (IaC), the IaU F-score is ``borderline''
statistically significantly better (11\% significance level). The IaU
F-score is statistically significantly better than the F-scores with
either using \newcite{BVD99} (IaB) or not using any initial
annotations (NI).

As expected, most (42 of 48) of the overall increase in recall
going from NI to IaU comes from increasing the recall of the {\em
object}, {\em time modifier} and other {\em modifier} relations, the
relations that IaC and IaB concentrate on. The ER for {\em object} is
11\% and for {\em time modifier} is 56\%.

When this combining approach is repeated the full
1963 GR instance training set, the overall results for the test set are
\begin{center}
\begin{tabular}{|l|r|r|r|r|}\hline
\multicolumn{5}{|c|}{Full Training Set, Overall Results}\\
 & R & P & F & ER \\ \hline
IaU & {\em 502 (67.1\%)} & 77.7\% & {\em 72.0\%} & 7.3\%\\ \hline
IaC & {\em 487 (65.1\%)} & {\em 79.7\%} & {\em 71.7\%} & 6.3\%\\ \hline
IaB & 486 (65.0\%) & 76.5\% & 70.3\% & 1.7\% \\ \hline
NI  &  476 (63.6\%) & 77.3\% & 69.8\% &\\ \hline
\end{tabular}
\end{center}
Compared to the smaller training set results, the difference between
IaU and IaC here is smaller for both the absolute F-score (0.3\%
versus 1.1\%) and ER (1.0\% versus 3.3\%). In fact, the F-score
difference is small enough to not be statistically significant. Given
the previous results for IaC and IaB as a small training set gets
larger, this is not surprising.

\section{Discussion}

GRs are important, but different sets of GRs are useful for different
purposes and different systems are better at finding certain types of
GRs. Here, we have been looking at ways of improving automatic GR
finders when one has only a small amount of data with the desired GR
annotations. In this paper, we improve the performance of the
\newcite{FVY99} GR transformation rule learner by using existing
systems to find related sets of GRs. The output of these systems is
used to supply initial sets of annotations for the rule learner. We
achieve modest gains with the existing systems tried. When one
examines the results, one notices that the gains tend to be uneven,
with a few GR types having large gains, and the rest not being
affected much. The different systems concentrate on improving
different GR types. We leverage this tendency to make a further modest
improvement in the overall results by providing the rule learner with
the merged output of these existing systems. We have yet to try other
ways of combining the output of existing systems that do not require
extra training data. One possibility is the example-based combiner in
\newcite[Sec.~3.2]{BandW98}.\footnote{Based on the paper, we were
unsure if extra training data is needed for this combiner. One of the
authors, Wu, has told us that extra data is not needed.} Furthermore,
finding additional existing systems to add to the combination may
further improve the results.

%\bibliographystyle{acl}

%\bibliography{nl}

\end{document}